%% file: causal-effect-estimation-for-image-interventions.tex
\documentclass{article} 
\usepackage{aaai25}  % DO NOT CHANGE THIS
\usepackage{times}  % DO NOT CHANGE THIS
\usepackage{helvet}  % DO NOT CHANGE THIS
\usepackage{courier}  % DO NOT CHANGE THIS
\usepackage[hyphens]{url}  % DO NOT CHANGE THIS
\usepackage{graphicx} % DO NOT CHANGE THIS
\usepackage{natbib}  % DO NOT CHANGE THIS AND DO NOT ADD ANY OPTIONS TO IT
\usepackage{caption} % DO NOT CHANGE THIS AND DO NOT ADD ANY OPTIONS TO IT
\usepackage{algorithm}
\usepackage{algorithmic}
\usepackage{mathtools}

\usepackage{newfloat}
\usepackage{listings}
\usepackage{booktabs}       % professional-quality tables
\usepackage{amsfonts}       % blackboard math symbols
\usepackage{nicefrac}       % compact symbols for 1/2, etc.
\usepackage{microtype}      % microtypography
\usepackage{xcolor}         % colors
\usepackage{amsmath, amsthm}
\usepackage{multirow}
\usepackage{adjustbox, lipsum}
\usepackage{multirow}
\usepackage{colortbl}
\usepackage{caption}

\usepackage{tikz}
\usepackage{tikzpeople}
\usetikzlibrary{fit,arrows.meta,mindmap,shadows,backgrounds,calc,positioning,shapes.misc}
\usepackage{xcolor}
\usetikzlibrary{arrows,3d,calc}
\usepackage{tikz-3dplot}
\usetikzlibrary{shadings, shapes.geometric, arrows.meta}
\usepackage{graphicx}
\newcommand{\imagepath}{Balto.jpg}
\newcommand{\imgpathone}{Balto.jpg}
\newcommand{\imgpathtwo}{The_Two_Mouseketeers.jpg}
\newcommand{\imgpaththree}{High_School.jpg}
\newcommand{\imgpathfour}{Dora_and_the_Lost_City_of_Gold.jpg}
\newcommand{\imgpathfive}{Bella.jpg}
\newcommand{\imgpathsix}{Bears.jpg}

\DeclareCaptionStyle{ruled}{labelfont=normalfont,labelsep=colon,strut=off} % DO NOT CHANGE THIS
\floatstyle{ruled}
\newfloat{listing}{tb}{lst}{}
\floatname{listing}{Listing}
\title{I See, Therefore I Do: Estimating Causal Effects for Image Treatments}
\author{
    Abhinav Thorat\equalcontrib,
    Ravi Kolla\equalcontrib,
    Niranjan Pedanekar
}
\affiliations{
    Sony Research India\\
    abhinav.thorat@sony.com, ravi.kolla@sony.com, niranjan.pedanekar@sony.com
}

\usepackage{bibentry}

\newtheorem{assumption}{Assumption}

\begin{document}
\maketitle
\begin{abstract}
Causal effect estimation under observational studies is challenging due to the lack of ground truth data and treatment assignment bias. Though various methods exist in literature for addressing this problem, most of them ignore multi-dimensional treatment information by considering it as scalar, either continuous or discrete. Recently, certain works have demonstrated the utility of this rich yet complex treatment information into the estimation process, resulting in better causal effect estimation. However, these works have been demonstrated on either graphs or textual treatments. There is a notable gap in existing literature in addressing higher dimensional data such as images that has a wide variety of applications. In this work, we propose a model named \textbf{NICE} (\textbf{N}etwork for \textbf{I}mage treatments \textbf{C}ausal effect \textbf{E}stimation), for estimating individual causal effects when treatments are images. NICE demonstrates an effective way to use the rich multidimensional information present in image treatments that helps in obtaining improved causal effect estimates. To evaluate the performance of NICE, we propose a novel semi-synthetic data simulation framework that generates potential outcomes when images serve as treatments. Empirical results on these datasets, under various setups including the \textit{zero-shot} case, demonstrate that NICE significantly outperforms existing models that incorporate treatment information for causal effect estimation.
\end{abstract}

\input{introduction}

\input{literature-survey}

\input{problem-formulation}

\input{proposed-model}

\input{data-simulation}

\input{experiments}

\input{conclusion}

\newpage
\bibliography{references}
\end{document}

%% file: introduction.tex
\section{Introduction}
\label{sec:introduction}
Causal effect estimation under observational studies is a critical yet challenging problem in the realm of causal inference as it enables a clear understanding of the impact of specific treatments or interventions on particular outcomes. It has omnipresent applications including, but not limited to, domains such as healthcare, economics, social sciences, education, entertainment and e-commerce. In this work, we specifically study \textbf{I}ndividual \textbf{T}reatment \textbf{E}ffect (ITE) estimation which is the most granular causal effect estimation task. ITE focuses on estimating the impact of a treatment at an individual level, as opposed to average causal effects, which apply to entire populations or sub-populations. ITE helps to achieve personalization of treatments based on user attributes. A few use cases of ITE across various domains include,  personalization of content, product, and investment plan recommendations in the entertainment, e-commerce, and finance industries, respectively.

In causal effects estimation literature, majority of the works represent treatments in one-hot encoding format or categorical nature. But, often these treatments are multi-dimensional such as images and graphs, do contain rich information and hence potentially can be used in the causal effects estimation if made available. This raises a question whether the causal effects estimates can be improved, in particular ITEs, by utilizing treatment attributes in the estimation process. To that end, there are a limited number of works~\citep{harada2021graphite, kaddour2021causal, nilforoshan2023zero} in the literature that try to address the above question. These works demonstrated ways of effective utilization of treatment attributes in the ITE estimation and showcased improved results. However, all these works have considered either graph or textual treatments, in their respective experiments. In this work, we study the ITE estimation problem for treatments being images and demonstrate an effective way to utilize treatment attributes that results into improved ITE estimates.   

We provide motivation for our problem, ITE estimation for image treatments, with the following prevalent use cases in our daily lives.
Consider an OTT (Over-The-Top) or video hosting application that displays its contents using thumbnails to the users. Suppose each content is present in multiple thumbnail variants and the goal is to personalize thumbnails for users to maximize the user engagement. This problem can be posed as an ITE estimation problem where thumbnails and users preferences to them correspond to treatments and their effects respectively. 

Consider another application in e-commerce or website publishing that sells products by displaying product images on their platform. Each product typically features multiple images (photos) captured from various angles and under different lighting conditions. Suppose, if the goal is to personalize product display images to maximize click through rates then it can be approached using our framework by considering product images as treatments and estimating users' preferences as treatment effects. 

We now briefly talk about the key challenges in our work. First and foremost challenge is the lack of existing datasets for the ITE estimation of image treatments. It necessitated us to simulate a new dataset that required extensive research and experimentation in terms of the appropriate image data and mathematical formulation of their induced effects on users. Second, to the best of our knowledge, there are very few existing works~\citep{harada2021graphite, kaddour2021causal, nilforoshan2023zero} that leverage treatment information in the literature, with no empirical results specifically addressing image treatments in ITE estimation.This gap presents a significant challenge for addressing the task at hand. Finally, the inheritance nature of our setup containing multiple treatments under observational studies increases the problem complexity in terms of confounding bias. We outline the salient contributions of our work that address the aforementioned challenges. 
\begin{itemize}
\item We propose a semi-synthetic data simulation setup that generates potential outcomes for the case of multiple image treatments.
\item We propose a novel neural network architecture, NICE, for ITE estimation of image treatments with a combination of \textbf{M}ean \textbf{S}quare \textbf{E}rror (MSE) and \textbf{M}aximum \textbf{M}ean \textbf{D}iscrepancy (MMD) losses.
\item We showcase the superior performance of NICE against baselines, on the Rooted \textbf{P}recision in the \textbf{E}stimation of \textbf{H}eterogeneous \textbf{E}ffects (PEHE) metric, across various treatment assignment bias conditions in numerical experiments.
\item We also conduct experiments under zero-shot scenarios where models are evaluated on unseen treatments during the training. Under these settings too, we observe that NICE outperforms baselines by a significant margin.  
\end{itemize}

We organize the rest of the paper as follows. We provide a comprehensive survey of related work in the following section. We introduce the mathematical formulation of the problem in the Problem Formulation section. The Proposed Model section then provides the technical details of our proposed NICE model. The subsequent sections, Data Simulation and Experiments, detail our data simulation setup and present a comparative analysis of our numerical results. Finally, we conclude the work and outline few potential future research directions in the Conclusions section.

%% file: literature-survey.tex
\section{Literature Survey}
\label{sec:literature-survey}
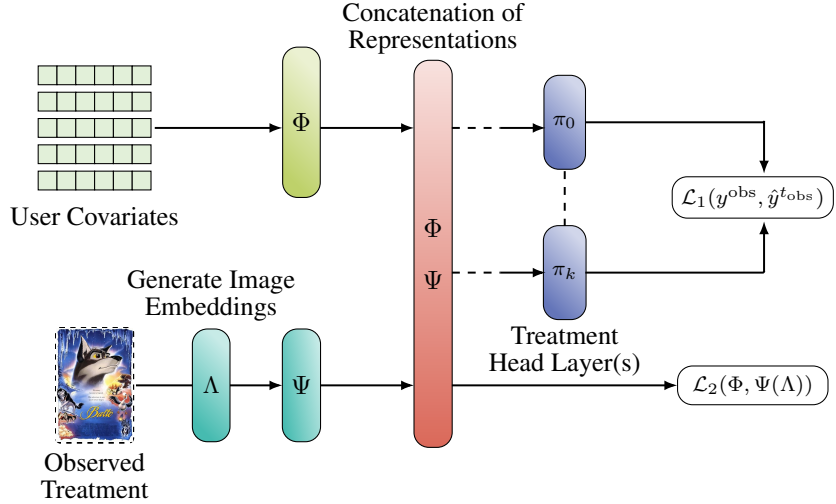
\begin{figure*}[!ht]
\centering
\begin{tikzpicture}
    \definecolor{lightgray}{HTML}{E2F0D9}
    \definecolor{arrow}{HTML}{000000}
    \definecolor{back}{HTML}{1952A6}
    \definecolor{gcn}{HTML}{BCBEDC}
    \definecolor{phi}{HTML}{DAF467}
    \definecolor{treat}{HTML}{FBA39D}
 	\definecolor{l2}{HTML}{B1D4FC}
 	\definecolor{cov}{HTML}{BED16A}
    \definecolor{obst}{HTML}{6D70D1}
    \definecolor{resnet}{HTML}{46BBB0}
    \definecolor{concat}{HTML}{DA6758}
    \definecolor{reg}{HTML}{6475B9}
    \definecolor{l1}{HTML}{FFFFFF}

  \pgfmathsetmacro\squareSize{0.25}
    \pgfmathsetmacro\startX{0.25}
    \pgfmathsetmacro\startY{-1.0}
    
    \foreach \x in {1,2,3,4,5,6} {
        \draw[fill=lightgray] (\startX + \x*\squareSize, \startY) rectangle ++(\squareSize, \squareSize);
    }
    
    \pgfmathsetmacro\squareSize{0.25}
    \pgfmathsetmacro\startX{0.25}
    \pgfmathsetmacro\startY{-1.35}
    
    \foreach \x in {1,2,3,4,5,6} {
        \draw[fill=lightgray] (\startX + \x*\squareSize, \startY) rectangle ++(\squareSize, \squareSize);
    }
    
    \pgfmathsetmacro\squareSize{0.25}
    \pgfmathsetmacro\startX{0.25}
    \pgfmathsetmacro\startY{-1.7}
    
    \foreach \x in {1,2,3,4,5,6} {
        \draw[fill=lightgray] (\startX + \x*\squareSize, \startY) rectangle ++(\squareSize, \squareSize);
    }
    
    \pgfmathsetmacro\squareSize{0.25}
    \pgfmathsetmacro\startX{0.25}
    \pgfmathsetmacro\startY{-2.05}
    
    \foreach \x in {1,2,3,4,5,6} {
        \draw[fill=lightgray] (\startX + \x*\squareSize, \startY) rectangle ++(\squareSize, \squareSize);
    }
	
	\pgfmathsetmacro\squareSize{0.25}
    \pgfmathsetmacro\startX{0.25}
    \pgfmathsetmacro\startY{-2.4}
    
    \foreach \x in {1,2,3,4,5,6} {
        \draw[fill=lightgray] (\startX + \x*\squareSize, \startY) rectangle ++(\squareSize, \squareSize);
    }

    % First Covariate text
    \node at (1.25, -2.75) [align=center,font=\fontsize{7}{8}] {User Covariates};
    
    % First Observed treatment block (with image and dashed border)
    \node[rectangle, 
          draw, 
          dash pattern=on 2pt off 2pt,
          line width=1.5pt,
          rounded corners=0.5mm,
          minimum width=0.5cm, 
          minimum height=1.5cm,
          inner sep=0pt] 
    at (1.25, -5) {\includegraphics[width=1cm, height=1.5cm]{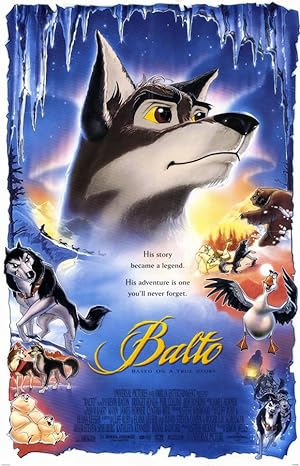}};

    % First Observed Treatment Text
    \node at (1.25, -6.2) [align=center,font=\fontsize{7}{8}] {Observed\\[-0.15ex] Treatment};
    
     % ResNet Embedding block (vertical elongated trapezium)
   \node[draw, shading=axis,
          left color=resnet,
          right color=resnet!30!white,
          shading angle=135, 
          draw,rounded corners=2mm, fill=resnet, minimum width=0.5cm, 
          minimum height=1.5cm,,font=\fontsize{9}{10}] at (2.8,-5) {$\Lambda$};
          
     % Second Observed Treatment Text
    \node at (2.8,-3.8) [align=center,font=\fontsize{6}{7}] {Generate Image \\[-0.25ex] Embeddings};
%
%    % ResNet Embedding text
%    \node at (4, -6.5) [font=\fontsize{6}{8}] {$\T_{obs}$ \\Representation};
    
    % Second Covariate block
    \node[rectangle, 
          shading=axis,
          left color=cov,
          right color=cov!30!white,
          shading angle=135, 
          draw, 
          rounded corners=2mm,
          minimum width=0.5cm, 
          minimum height=2cm,
          font=\fontsize{10}{12}] 
    at (4,-1.5) {$\Phi$};

    % Second Covariate text
    \node at (4, -2.6) [align=center,font=\fontsize{8}{10}] {};
    
    % Second Observed treatment block
    \node[rectangle,
          shading=axis,
          left color=resnet,
          right color=resnet!30!white,
          shading angle=100, 
          draw, 
          rounded corners=2mm, 
          minimum width=0.5cm, 
          minimum height=1.5cm,
          font=\fontsize{10}{12}] 
    at (4,-5) {$\Psi$};

    % Second Observed Treatment Text
    \node at (4, -6.15) [align=center,font=\fontsize{8}{10}] {};
    
    % Second Observed treatment block
    \node[rectangle,
          shading=axis,
          left color=concat,
          right color=concat!20!white,
          shading angle=180, 
          draw, 
          rounded corners=2mm, 
          minimum width=0.5cm, 
          minimum height=5.15cm,
          font=\fontsize{10}{12},
          align=center] 
    at (5.75,-3.25) {$\Phi$ \\[-0.5ex]\\$\Psi$ };

    % Second Observed Treatment Text
    \node at (5.75, -0.25) [align=center,font=\fontsize{7}{8}] {Concatenation of \\[-0.25ex] Representations};

	%Lines for First part if network
	
	% line from cov to phi 
	\draw[-{Latex[length=1.75mm]}, line width=0.75pt, arrow] (2.05,-1.575) -- (3.75,-1.575);
    
    % line from Obs t to resnet 
	%\draw[-{Latex[length=2.5mm]}, line width=1pt, arrow] (1.8,-5) -- (2.55,-5);
	\draw[-,line width=0.75pt, arrow] (1.8,-5) -- (2.55,-5);
	% line from resnet to psi
	\draw[-{Latex[length=1.75mm]}, line width=0.75pt, arrow] (3.05,-5) -- (3.75,-5);
	% line from phi to concat
	\draw[-{Latex[length=1.75mm]}, line width=0.75pt, arrow] (4.25,-1.575) -- (5.5,-1.575);
	% line from psi to concat
	\draw[-{Latex[length=1.75mm]}, line width=0.75pt, arrow] (4.25,-5) -- (5.5,-5);
    
    % Head layers first block
    \node[rectangle, 
          shading=axis,
          left color=reg,
          right color=reg!30!white,
          shading angle=135, 
          draw, 
          rounded corners=2mm,
          minimum width=0.5cm, 
          minimum height=1.25cm,
          font=\fontsize{8}{10}] 
    at (7.5,-1.5) {$\pi_0$};
    
    % Head layers second block
    \node[rectangle, 
          shading=axis,
          left color=reg,
          right color=reg!30!white,
          shading angle=135, 
          draw, 
          rounded corners=2mm,
          minimum width=0.5cm, 
          minimum height=1.25cm,
          font=\fontsize{8}{10}] 
    at (7.5,-3.5) {$\pi_k$};
    
    % Second Observed Treatment Text
    \node at (7.5, -4.55) [align=center,font=\fontsize{7}{8}] {Treatment \\[-0.15ex]Head Layer(s)};
    
    %Lines from concat to pi 1  
    \draw[-,dashed, line width=0.75pt, arrow] (6,-1.575) -- (6.65,-1.575);
     \draw[-{Latex[length=1.75mm]}, line width=0.75pt, arrow] (6.65,-1.575) -- (7.25,-1.575);
     
     %Lines from concat to pi k  
    \draw[-,dashed, line width=0.75pt, arrow] (6,-3.5) -- (6.65,-3.5);
     \draw[-{Latex[length=1.75mm]}, line width=0.75pt, arrow] (6.65,-3.5) -- (7.25,-3.5);
     
     % L1 loss block
    \node[rectangle, 
          shading=axis,
          left color=l1,
          right color=l1!30!white,
          shading angle=135, 
          draw, 
          rounded corners=2mm,
          minimum width=1.5cm, 
          minimum height=0.5cm,
          font=\fontsize{8}{10}] 
    at (10,-2.5) {$\mathcal{L}_1(y^{\mathrm{obs}}, \hat{y}^{t_{\mathrm{obs}}})$};
    
    % L2 loss block
    \node[rectangle, 
          shading=axis,
          left color=l1,
          right color=l1!30!white,
          shading angle=135, 
          draw, 
          rounded corners=2mm,
          minimum width=2cm, 
          minimum height=0.5cm,
          font=\fontsize{8}{10}] 
    at (10,-5) {\(\mathcal{L}_2(\Phi, \Psi(\Lambda))\)};
    
    %line concat to l2
    \draw[-{Latex[length=1.75mm]}, line width=0.75pt, arrow] (6,-5) -- (9,-5);
    
    %%% line from pi 0 to l1
    % Define nodes
    \node (P) at (7.65, -3.5){} ;
    \node (Q) at (10.15, -2.69) {};

    % Draw the line with an upward curve and arrow
    \draw[-{Latex[length=1.75mm]}, line width=0.75pt, rounded corners=0.5pt] 
        (P) -- ++(2.5, 0) |- ++(0, 0) -- (Q);

    %%% line from pi k to l1
    % Define nodes
    \node (R) at (7.65, -1.5){} ;
    \node (S) at (10.15, -2.31) {};

    % Draw the line with an upward curve and arrow
    \draw[-{Latex[length=1.75mm]}, line width=0.75pt, rounded corners=0.5pt] 
        (R) -- ++(2.5, 0) |- ++(0, 0) -- (S);

   % Architecture title with boldface only for abbreviation letters
    \node at (5.75, -7) [
        align=center,
        font=\fontsize{10}{12}\normalfont
    ] {
       %\textbf{NICE} : \textbf{N}etwork for \textbf{I}mage treatments \textbf{C}ausal effect \textbf{E}stimation
    };

    %vertical dashed line between head network
    \draw[-,dashed, line width=0.75pt, arrow] (7.5,-2.15) -- (7.5,-2.88);
	
 \end{tikzpicture}
 \label{fig:workflow-overview}
\caption{\textbf{NICE} : \textbf{N}etwork for \textbf{I}mage treatments \textbf{C}ausal effect \textbf{E}stimation}
 \end{figure*}

In this work, we deal with the estimation of ITEs that hold at individual unit level as opposed to the predominantly studied Average Treatment Effects (ATE)~\citep{shpitser2012identification, pearl2017detecting}, which hold at whole population level, estimation in the literature. Specifically, we study ITE estimation under multiple image treatments setup by utilizing treatments information in the estimation hence we restrict ourselves contrasting our work with only ITE estimation under multiple treatments and (or) works that utilized treatments information in the estimation. Most of the works that involve multiple treatments~\citep{yoon2018ganite, schwab2018perfect, guo2020learning, schwab2020learning, thorat2023estimation} do not consider the rich multi-dimensional treatment information in the estimation and merely represent them as scalars using one hot encoding. 

There are few existing works~\citep{harada2021graphite, kaddour2021causal, nilforoshan2023zero} that utilized treatments information in the ITE estimation under multiple treatments setup. The authors in~\citep{harada2021graphite} consider the problem of ITE estimation for multiple graph treatments, different from our setup of image treatments, and demonstrated ways to utilize the graph treatments information to obtain improved causal effects estimates. The work in~\citep{kaddour2021causal} deals with the ITE estimation for structured treatments such as graph, image and textual by incorporating treatments information in the estimation process. However, their proposed algorithm, SIN, is evaluated only on the datasets with graph treatments in the their experiments. Hence, SIN's performance on image treatments datasets has not been studied yet, making SIN one of the baselines in our work. The following latest work~\citep{nilforoshan2023zero} also utilizes treatments information for ITE estimation but it is primarily on the zero-shot tasks, estimation of treatment effects on the unseen treatments by model during training.       

%% file: problem-formulation.tex
\section{Problem Formulation}
\label{sec:problem-formulation}
In this section, we briefly outline the problem considered in this work. 
Let $k \in \mathbb{N}$ and $n \in \mathbb{N}$ denote the number of available treatments and the number of instances/users. We use $i$, $\mathbf{x}$ and $t$ for referencing users, their covariates and treatments respectively. Let $\mathbf{x}_i \in \mathcal{X} \subset \mathbb{R}^d$, $t_i \in \lbrace 1, 2, \cdots , k \rbrace$, denote covariates, index of assigned treatment of user-$i$. We use $I_t \in \mathcal{I} \subset \mathbb{R}^m$ to denote the image corresponding to the treatment-$t.$

We follow the Rubin-Neyman~\citep{rubin2005causal} potential outcomes framework for introducing the problem. Let $y_i^{t}$ denotes the potential outcome of user-$i$ when treatment-$t$ is applied. Since $t_i$ is the treatment given to user-$i$, $y_i^{t_i}$ denotes the observed outcome or factual outcome of user-$i.$ For brevity purposes, we write $y_i^{t_i}$ as $y_i^{t}$. Given user-$i$ with covariates $\mathbf{x}_i$, and a pair of treatments $a$, $b$, let us define ITE of treatment $a$ w.r.to $b$, using notation $\tau^{a, b}(\mathbf{x}_i)$ , as below:
\begin{equation}
\label{eq:ITE-equation}
\tau^{a, b}(x_i) = \mathbb{E} \left[ y_i^a \mid \mathbf{x}=\mathbf{x}_i \right] - \mathbb{E} \left[ y_i^b \mid \mathbf{x}=\mathbf{x}_i \right].
\end{equation}

In our setup, we assume that the treatments are images and are available to the model. In the following, we provide the technical formulation of our problem statement. Given $n$ observations, $ \lbrace \mathbf{x}_i, I_{t_i} , y_i^{t_i} \rbrace_{i=1}^n$, of users with covariates $\mathbf{x}_i$, their assigned treatment images, $I_{t_i}$ and the corresponding observed potential outcomes, $y_i^{t_i}$ our goal is to estimate ITEs, given in equation~\eqref{eq:ITE-equation}, for all pairs of treatments.

For the quantification of a model’s performance we use the standard metric in the literature named PEHE whose formulation is given below:
\begin{equation}
\label{eq:epsilon-pehe-formula} 
\epsilon_{\text{PEHE}}  = \frac{1}{{k \choose 2}} \sum\limits_{a=1}^{k} \sum\limits_{b=0}^{a-1}\left[ \frac{1}{n} \sum\limits_{i=1}^n (\hat{\tau}^{a, b}(\mathbf{x}_i) - \tau^{a, b}(\mathbf{x}_i))^2 \right], 
\end{equation}
where $\hat{\tau}\left( \cdot \right)$ represents the estimated PEHEs produced by a model.

%% file: proposed-model.tex
\section{Proposed Model}
The NICE framework, designed for ITE estimation with image-based treatments, is founded on the principle of strong ignorability~\citep{rubin2005causal}. This principle encompasses two essential assumptions: conditional independence (unconfoundedness) and positivity, both formally articulated below.
\begin{assumption}[Conditional Independence (Unconfoundedness)]
\label{def:ConditionalIndependence}
Given the covariates, $\mathbf{x}_i$, the treatment assignment, $t,$ is independent of the potential outcomes, ensuring that all confounding variables affecting both treatment assignment and outcomes are accounted for within $\mathbf{x}_i$:
\begin{equation*}
(y^1_i, y^2_i,...y^k_i) \perp t \mid \mathbf{x}_i.
\end{equation*}
\end{assumption}

\begin{assumption}[Positivity (Overlap)]
\label{def:Positivity}
This assumption ensures that every subgroup defined by $ \{ \mathbf{x}=\mathbf{x}_i \}$ has a positive probability of receiving any of the treatments $a \in \{ 1, 2, \cdots, k \}.$ It guarantees sufficient variation in treatment assignments observed across all values of $\mathcal{X}$, which can be expressed as:
\begin{equation*}
0 < P(t = a \mid \mathbf{x} = \mathbf{x}_i) < 1 \, \, \forall 1 \leq a \leq k.
\end{equation*}
\end{assumption}

Estimating causal effects is challenging because some confounders are difficult to measure directly in the observational data. This limitation undermines the strong ignorability condition, which presumes that all confounders are accounted for in the data. The absence of this condition may make the causal effect estimates biased, as unobserved confounders can influence both treatment assignment and outcomes.

The challenge of confounding is further exacerbated for the case of image treatments. Due to their complex and detailed nature, images can introduce additional layers of variability that are not always directly observable. By incorporating these rich treatment attributes, ITE estimation can account for hidden confounders, thereby improving the accuracy of ITE estimates.

Our proposed model, NICE, addresses ITE estimation by utilizing treatment information, specifically images. Figure~1 illustrates the detailed architecture of the NICE model, comprising three key steps mentioned below.
\begin{itemize}
\item[A.] Generating representations for both covariates and treatments simultaneously and then concatenating these outputs
\item[B.] Employing individual treatment head networks to generate counterfactual estimates
\item[C.]Computing a treatment regularization loss to mitigate the treatment assignment bias or confounding bias along with regression loss to ensure
accurate predictions.
\end{itemize} 
This structured approach enables our model to effectively account for the complex information contained in images, thereby improving the accuracy of counterfactual estimations. Detailed explanations of our model architecture are as follows.

\subsection{A. Learning Representation of User Covariates and Observed Image Treatments}
We employ two distinct fully connected networks to learn representations of covariates, $\mathbf{x} \in \mathcal{X},$ and observed image treatments $I_t \in \mathcal{I}$, capturing their low-dimensional embeddings. Specifically, we define two functions, $\Phi: \mathcal{X} \rightarrow \mathbb{R}^{d_1}$ and $\Psi: \Lambda (\mathcal{I}) \rightarrow \mathbb{R}^{d_2},$ to extract representations for covariates and treatment images. 

The utility of learning covariate representations to enhance causal effect estimation has been previously demonstrated in the literature~\citep{shalit2017estimating}. Similarly, learning treatment representations has been explored, particularly in graph-based contexts, as in Graphite\citep{harada2021graphite}, SIN~\citep{kaddour2021causal} and CaML~\citep{nilforoshan2023zero} algorithms. In our approach, we first use an existing image embedding model, denoted by $\Lambda,$ for obtaining image embeddings. Then, these image embeddings are fed to a representation network, $\Psi.$ Currently, we considered two popular well studied models in the literature such as ResNet~\citep{he2016deep} and VGG~\citep{simonyan2014very} as candidates for $\Lambda$ in the NICE model. In particular, $\Lambda$ is used solely to infer image treatment embeddings and is not a trainable component in the NICE model.   
%In our approach, we derive embeddings of observed treatment images by employing the ResNet \citep{he2016deep} and VGG\citep{simonyan2014very} models, further improving the embeddings with the mapping function \(\Psi\) to adapt for causal effect estimation. 
%The mathematical notations for covariates and treatments representation are given by \(\Phi(\mathbf{x})\)$\Psi(\Lambda(I_{t_{obs}})).$ 

Note that, representation of $\Psi$ as a one-hot encoding of discrete treatments aligns with the standard multiple treatments setting. However, this approach fails to leverage the rich structural information inherent to image treatments and consequently suffers in causal effect estimates.
 %when dealing with a large number of treatments. 
Further, we concatenate the covariates and treatment representations to create a joint embedding, which is then utilized by the treatment head networks for ITE estimation.

% Algorithm code
\begin{algorithm}
\caption{NICE Training}
\begin{algorithmic}
\STATE \textbf{Input:} Observational data: $ \mathcal{D} = \{ \left( \mathbf{x}_i, I_{t_i}, y^t_i \right) \}_{i=1}^n \sim \mathcal{D}_{\text{train}},\mathcal{D}_{\text{val}}$, and hyper parameters $\alpha \geq 0$ and $\beta \geq 0.$
\STATE \textbf{Output:} An outcome prediction model consisting of Treatment Head Networks: $f \rightarrow \pi_k = (\Phi(\mathbf{x}), \Psi(\Lambda (I_{t_{obs}}))$
\end{algorithmic}
\begin{algorithmic}[1]
\setcounter{ALC@line}{0}
\STATE Initialize parameters: $\Phi, \Psi, \pi_k$
\WHILE{{\textit{not converged}}}
    \STATE Sample a mini-batch  \\ $ B = \{ (x_{i_o}, I_{t_{i_o}}, y^{t_{i_o}}_{i_o}) \}_{o=1}^B \subset \mathcal{D}_{\text{train}} $
    \STATE Mini-batch approximation of Regression Loss \(\mathcal{L}_1\):
    \STATE Compute:
    \[
    \mathcal{L}_1 = \frac{1}{|B|} \sum_{o=1}^{|B|} (\hat{y}^{t_{i_o}}_{i_o} - y^{t_{i_o}}_{i_o})^2
    \]
    \STATE Mini-batch approximation of the Treatment Regularization Loss \(\mathcal{L}_2\):
    \[
    \mathcal{L}_2 = \frac{1}{\binom{k}{2}} \sum_{a=1}^{k} \sum_{b=1}^{a-1} 
\mathrm{MMD}\left( \lbrace \Phi; \Psi(\Lambda)) \rbrace_{t=a}, 
\lbrace \Phi; \Psi(\Lambda) \rbrace_{t=b} \right)
    \]
    \STATE \textbf{Update Functions}:
    \[
    f(\Phi,\Psi,\pi_k) \leftarrow f ( \Phi,\Psi,\pi_k) - \lambda.\nabla( f ({\Phi,\Psi,\pi_k}))
    \]
    \STATE \textbf{Minimize}:
    \[
    \alpha \cdot \mathcal{L}_1 + \beta \cdot \mathcal{L}_2
    \]
    using SGD
\ENDWHILE
\end{algorithmic}
\end{algorithm}

\subsection{B. Treatment Head Networks}
In the second part of our model, we leverage concatenated embeddings of user covariates and treatments representations as a unified input to distinct treatment head networks corresponding to each treatment category.  
Given $k$ available treatments, we train $k$ fully connected networks to learn the functions for each individual treatment, aimed at estimating the potential outcomes. We denote these treatment head networks as $\pi_t$ for $ t \in \{ 1, 2 \cdots, k\}.$
Mathematically, for an instance $i$ with covariates $\mathbf{x}_i$ and observed treatment $t_{\mathrm{obs}} = t$, $\pi_t$ is defined as:
\begin{equation*}
\begin{aligned}
\pi_t \left( \Phi(\mathbf{x}_i), \Psi(\Lambda (I_{t_{obs}})) \right) &= w_t \sigma\left(W_t^l \cdots \right.\\
&\quad \left.\sigma\left(W_t^1 \left(\Phi(\mathbf{x}_i), \Psi(\Lambda (I_{t_{obs}}))\right)\right)\right),
\end{aligned}
\end{equation*}
where $W_t^l$ and $w_t$ represent the weights of the $l$-th FC layer and the regression layer in the network head-$t$, respectively. The neural network bias terms follow the same rule and are omitted here for simplicity.

With both components of the model described, our model’s prediction of the potential outcome for treatment $t$ given instance $i$ is defined as:
\begin{equation}
\label{eq:y-pred}
\hat{y}_t^i = \pi_t(\Phi(\mathbf{x}_i), \Psi(\Lambda (I_{t}))
\end{equation}

\subsection{C. Loss Function}
We optimize our model using two primary loss functions: (i) Regression Loss and (ii) Treatment Regularization Loss, denoted as $\mathcal{L}_1$ and $\mathcal{L}_2,$ respectively. These loss terms are balanced using parameters $\alpha$ and $\beta$.

To achieve high predictive accuracy on observed outcomes, we employ the traditional mean square error loss. Given that each treatment group exhibits a unique distribution, optimizing the regression loss enables us to capture the approximate means for each treatment group. Specifically, we optimize the head network corresponding to the observed treatment~$t.$ The regression loss function $\mathcal{L}_1$ is defined as:

\begin{equation}
\label{eq:loss1}
\mathcal{L}_1 = \frac{1}{n} \sum_{i=1}^{n} (\hat{y}^{t}_i - y_i)^2
\end{equation}

We use the Treatment Regularization loss, denoted by $\mathcal{L}_2,$ computed using both covariates and treatments representation, to address treatment assignment bias~\textemdash~a crucial step in our model. As mentioned earlier, we concatenate the covariates and treatments representations to form a joint embedding. Considering that the latent spaces of covariates and treatments may differ, our goal is to achieve a balanced representation that accounts for treatment assignment bias. To that end, special cases of Integral Probability Metrics (IPM) have been utilized in the literature to obtain a balanced representation~\citep{johansson2016learning}. We employ a special case of IPM, specifically Maximum Mean Discrepancy (MMD) loss~\citep{sriperumbudur2012empirical}, to obtain a balanced representation of joint embedding across all treatments. In particular,
the treatment regularization loss computes the average MMD distance between the joint embeddings of covariates and image treatment representations across all treatment pair combinations, whose mathematical formulation is given below:
\begin{equation}
\label{eq:loss2}
\mathcal{L}_2 = \frac{1}{\binom{k}{2}} \sum_{a=1}^{k} \sum_{b=1}^{a-1} 
\mathrm{MMD}\left( \lbrace \Phi; \Psi(\Lambda)) \rbrace_{t=a}, 
\lbrace \Phi; \Psi(\Lambda) \rbrace_{t=b} \right),
\end{equation}
where the notation $\{ \Phi ; \Psi \}$ denotes the concatenation operation. 
This approach aims to optimize both covariate and treatment representations, thereby mitigating the confounding effects introduced by the complex nature of treatments, in this case, images. 

In summary, the total loss function $\mathcal{L}$ used to optimize the NICE framework comprises a regression loss term and an IPM loss term. The hyperparameters $\alpha$ and $\beta$ balance these two loss components to achieve optimal performance. The combined loss $\mathcal{L}$ is defined as:
\begin{equation}
\label{eq:total_loss}
\mathcal{L} = \alpha \cdot \mathcal{L}_1 + \beta \cdot \mathcal{L}_2.
\end{equation}

%% file: data-simulation.tex
\section{Data Simulation}
\label{sec:data-simulation}
\begin{figure}
\centering
\resizebox{!}{6cm}{
\begin{tikzpicture}
% Define the size of the images and the spacing
\def\imagewidth{2.5cm}
\def\imageheight{3.5cm}
\def\hspace{0.25cm} % Horizontal space between images
\def\vspace{3.75cm} % Vertical space between rows

% Create a 2x3 grid of images with specific width and height for each
\node[inner sep=0pt] 
    at (0, 0) {\includegraphics[width=\imagewidth, height=\imageheight]{\imgpathone}};
    
\node[inner sep=0pt] 
    at (\imagewidth + \hspace, 0) {\includegraphics[width=\imagewidth, height=\imageheight]{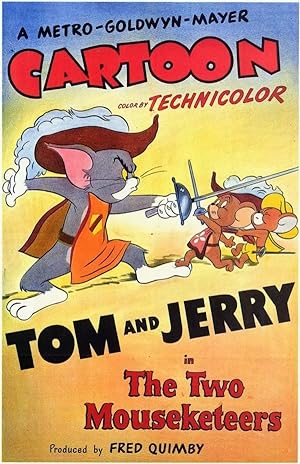}};
    
\node[inner sep=0pt] 
    at (2*\imagewidth + 2*\hspace, 0) {\includegraphics[width=\imagewidth, height=\imageheight]{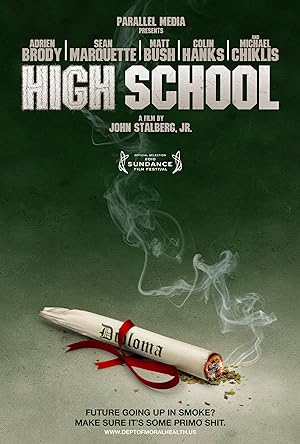}};
    
\node[inner sep=0pt] 
    at (0, -\vspace) {\includegraphics[width=\imagewidth, height=\imageheight]{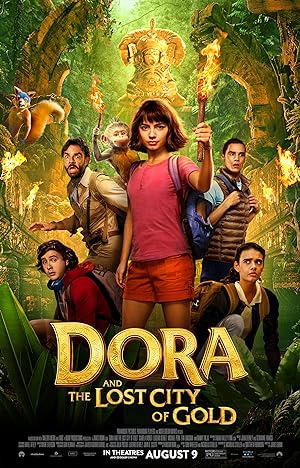}};
    
\node[inner sep=0pt] 
    at (\imagewidth + \hspace, -\vspace) {\includegraphics[width=\imagewidth, height=\imageheight]{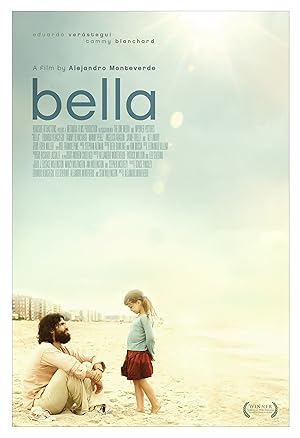}};
    
\node[inner sep=0pt] 
    at (2*\imagewidth + 2*\hspace, -\vspace) {\includegraphics[width=\imagewidth, height=\imageheight]{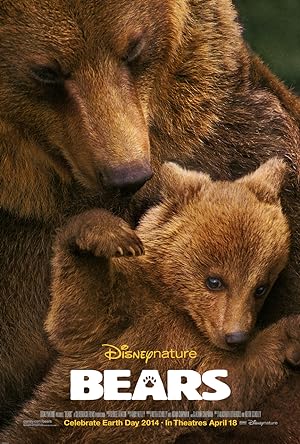}};
    
\end{tikzpicture}
}
\label{fig:sample-posters}
\caption{Few example posters considered as treatments}
\end{figure}

Evaluating causal effects presents inherent challenges, primarily due to the difficulty in obtaining ground truth for counterfactual outcomes in observational datasets. The existing literature often addresses this by synthetically generating potential outcomes resulting into semi-synthetic datasets. In other words, the covariates and treatments can correspond to real-world data, while the potential outcomes are synthetically generated. However, to the best of our knowledge, there are no existing datasets available for our problem that involve images as treatments. Therefore, we focus on generating new semi-synthetic datasets to evaluate NICE algorithm against the baselines. Specifically, in our case, treatment images correspond to real world data, user covariates and potential outcomes are synthetically generated. 
  
We now briefly outline the details of our datasets generation process. We generate our dataset using PosterLens~\citep{aptlin2021posterlens} dataset that contains posters of various movies and their respective ResNet embeddings. We first randomly draw 20k posters' ResNet~\citep{he2016deep} embeddings of size 512 from the PosterLens dataset. These embeddings are used as a proxy for users' covariates, $\mathbf{x},$ mentioned in the Problem Formulation section. For a given user with covariates, our goal is to estimate their opinion on the shown poster, a real valued scalar. Note that, here, treatments are the posters shown to users and their opinions are considered as a proxy for the treatment effects. Few example posters considered as treatments are in Figure~2. We generate multiple datasets based on the selected number of available treatments, which can be 4, 8 and 16. 

To generate potential outcomes for the $k$ number of treatments setup we first generate $(k+1)$ centroids as follows. Either randomly select $(k+1)$ ResNet embeddings from the 20,000 embeddings selected above or train a KMeans clustering algorithm on the 20,000 embeddings with $(k+1)$ clusters as an input and take the resultant centroids. We use $\textbf{z}_i$ to denote centroids. We use $y_{i}^t$ to denote final potential outcomes for user-$i$ and treatment-$t.$ Our final potential outcomes are product of two quantities $\tilde{y}_{i}^t$ and $d_{i}^t$ that are generated as follows. 
\begin{itemize}
\item Generation of $\tilde{y}_{i}^t$ : For each treatment-$t$, generate $\mu_t$ and $\sigma_t$ as follows: $\mu_t \sim \mathcal{N}(0.45, 0.15)$ and $\sigma_t \sim \mathcal{N}(0.1, 0.05)$. Then, $\tilde{y}_{i}^t$ is an i.i.d sample drawn from $\mathcal{N}(\mu_t, \sigma_t)$. Observe that $\tilde{y}_{i}^t$'s are stochastic in nature and the distribution solely dependent on the treatment.	
\item Generation of $d_{i}^t$ : It tries to measure the preference of user with covariates $\mathbf{x}_i$ to a treatment represented using its ResNet embedding which is defined as: \\
$d_{i}^t = \textbf{x}_i^T \textbf{z}_t + \textbf{x}_i^T \textbf{z}_{k+1}$  $\forall \, 1 \leq i \leq n \, \& \, 1 \leq t \leq k$.
\end{itemize}
Given the above, the final potential outcomes, denoted by $y_{i}^t$ for any $1 \leq i \leq n \, \& \, 1 \leq t \leq k,$ are defined as 
\begin{equation}
\label{eq:final-potential-outcomes}
y_{i}^t = c\tilde{y}_{i}^t d_i^t = c\tilde{y}_{i}^t \left[ \mathbf{x}_i^T \mathbf{z}_t + \mathbf{x}_i^T \mathbf{z}_{k+1} \right],
\end{equation}   
where $c>0$ is a fixed constant and we keep it as 5 in the experiments.

We now briefly explain the process of observed treatment generation. Let $p_{i, t}$ denote the probability of treatment-$t$ assigned to user-$i$, defined as below:
\begin{equation}
\label{eq:treatment-assignment-probability}
p_i^t = \frac{\exp \left(\kappa_i y_{i}^t \right)}{\sum\limits_{a=1}^k \exp \left( \kappa_i y_{i}^a \right)}, 
\end{equation}
where $\kappa = [\kappa_1, \kappa_2, \cdots, \kappa_k] > 0,$ is a set of parameters that controls the treatment assignment bias. In other words, choosing $\kappa_a \gg \kappa_b, \, b \neq a$ makes the treatment assignment distribution skewed towards $a^{\text{th}}$ treatment. For a given user-$i$, we randomly assign a treatment with the above probabilities, $p_{i}^t,$ and call it as the observed treatment for that user. 	

%% file: experiments.tex
\section{Experiments}
%%% PEHE table
\begin{table*}[!ht]
\centering
\caption{Performance comparison of NICE vs baslines on semi-synthetic datasets using images as treatments, with variations in the number of treatments. Here, $\kappa_a = 10,$ $ 1 \leq a \leq k$ is considered.}

\label{tab:pehe_comparison}
\begin{tabular}{lccc}
\hline
Method & $k=4$ & $k=8$ & $k=16$ \\
\hline 
%Naive & 139.33 $\pm$ 22.84 & 148.42  $\pm$ 29.7 & 155.04 $\pm$ 17.16 \\
TarNet & 128.25 $\pm$ 24.07 & 137.67 $\pm$ 26.75 & 152.77 $\pm$ 15.69 \\
GraphITE & 128.26 $\pm$ 24.52 & 135.33 $\pm$ 22.13 & 141.06 $\pm$ 12.76 \\
SIN & 127.74 $\pm$ 24.71 & 134.68 $\pm$ 22.06 & 139.36 $\pm$ 13.08 \\
CaML & 127.74 $\pm$ 24.71 & 138.0 $\pm$ 20.63 & 139.72 $\pm$ 13.05 \\
\rowcolor{gray!20}
\textbf{NICE - ResNet} & \textbf{91.95 $\pm$ 12.06} & \textbf{105.22 $\pm$ 14.92} & 114.38 $\pm$ 10.16 \\
\hline
\rowcolor{gray!20}
\textbf{NICE - VGG} & 98.95 $\pm$  13.16 & 107.57 $\pm$ 17.61 & \textbf{113.94 $\pm$ 9.97} \\
\hline
\end{tabular}
\end{table*}

%%% PEHE table zero shot 
\begin{table*}[!ht]
\centering
\caption{Performance comparison of NICE in a zero-shot setting against baselines, with variations in number of treatment to assess causal estimation on unseen treatments.  }

\label{tab:pehe_comparison_zero_shot}
\begin{tabular}{lccc}
\hline
Method & $k=4$ & $k=8$ & $k=16$ \\
\hline  
TarNet & 131.78 $\pm$ 29.23 & 150.71 $\pm$ 22.46 & 154.96 $\pm$ 23.68\\
GraphITE & 128.35 $\pm$ 25.24 & 136.68 $\pm$ 19.29 & 145.13 $\pm$ 23.59 \\
SIN     &   128.30 $\pm$ 26.54 & 133.20 $\pm$ 19.44 & 137.87 $\pm$ 13.90 \\
CaML & 120.68 $\pm$ 24.55 & 133.38 $\pm$  17.84 & 135.89 $\pm$ 13.97 \\
\rowcolor{gray!20}
\textbf{NICE - ResNet} & \textbf{92.74 $\pm$ 15.64} & \textbf{103.82 $\pm$10.79} &117.78$\pm$16.82 \\
\hline
\rowcolor{gray!20}
\textbf{NICE - VGG} & 98.61 $\pm$ 17.22& 106.79 $\pm$13.09 & \textbf{117.37 $\pm$17.77} \\
\hline
\end{tabular}
\end{table*}

%%% PEHE table kappa 50
\begin{table*}[!ht]
\centering
\caption{Performance comparison of NICE against baselines under varying treatment assignment bias $\kappa_a = 10, \, 1 \leq a < k$ and $\kappa_k= 50$, introducing skewness in treatment assignment to better reflect real-world scenarios. }

\label{tab:pehe_comparison_kappa_50}
\begin{tabular}{lccc}
\hline
Method & $k=4$ & $k=8$ & $k=16$ \\
\hline
%Naive & 139.33 $\pm$ 22.84 & 148.42  $\pm$ 29.7 & 155.04 $\pm$ 17.16 \\
TarNet & 136.18 $\pm$ 31.84 & 150.94 $\pm$ 27.51 & 152.85 $\pm$ 18.92 \\
GraphITE & 129.57 $\pm$ 24.63 & 141.33 $\pm$ 22.76 & 145.8 $\pm$ 13.71 \\
SIN & 127.74 $\pm$ 24.72 & 134.68 $\pm$ 22.06  & 139.66 $\pm$ 12.85 \\
CaML & 129.11 $\pm$ 23.96 & 137.3 $\pm$ 21.41 & 139.35 $\pm$ ± 13.87 \\
\rowcolor{gray!20}
\textbf{NICE - ResNet} & \textbf{91.71 $\pm$ 11.85} & \textbf{107.09 $\pm$ 16.92} & 128.65 $\pm$ 22.25 \\
\hline
\rowcolor{gray!20}
\textbf{NICE - VGG} & 97.39 $\pm$ 12.98 & 110.45 $\pm$ 18.44 & \textbf{128.51 $\pm$ 21.91} \\
\hline
\end{tabular}
\end{table*}

%%% PEHE table kappa 100
\begin{table*}[!ht]
\centering
\caption{Performance comparison of NICE against baselines under varying treatment assignment bias $\kappa_a = 10, \, 1 \leq a < k$ and $\kappa_k= 100$, introducing skewness in treatment assignment to better reflect real-world scenarios.}

\label{tab:pehe_comparison_kappa_100}
\begin{tabular}{lccc}
\hline
Method & $k=4$ & $k=8$ & $k=16$ \\
\hline
%Naive & 131.59 $\pm$ 22.63 & 146.45 $\pm$ 26.45 & 155.67 $\pm$ 18.83 \\
TarNet & 131.9 $\pm$ 25.79 & 139.65 $\pm$ 22.69 & 148.98 $\pm$ 18.57\\
GraphITE & 129.28 $\pm$ 24.33 & 142.99 $\pm$ 21.05 & 148.79 $\pm$ 13.15 \\
SIN          & 127.74 $\pm$ 24.72 & 134.68 $\pm$ 22.06 & 139.55 $\pm$ 12.74 \\
CaML & 127.74 $\pm$ 24.71 & 134.24 $\pm$ 21.83 & 135.39  $\pm$ 8.45 \\
\rowcolor{gray!20}
\textbf{NICE - ResNet} & 103.49 $\pm$ 41.21 & \textbf{107.49 $\pm$18.35} &\textbf{129.59 $\pm$23.84} \\
\hline
\rowcolor{gray!20}
\textbf{NICE - VGG} & \textbf{96.08 $\pm$ 13.16}& 108.6 $\pm$ 18.2 & 131.93$\pm$19.58 \\
\hline
\end{tabular}
\end{table*}

%% Hyperparam table
\begin{table}[h!]
\centering
\begin{tabular}{c|c}
\hline
\textbf{Hyperparameter} & \textbf{Search Range} \\
\hline
Number of layers ($\Phi$) & \{4, 6, 8\} \\
\hline
Number of nodes ($\Phi$) & \{200, 400, 600\} \\
\hline
Number of layers ($\Psi$) & \{4, 6, 8\} \\
\hline
Number of nodes ($\Psi$) & \{200, 400, 600\} \\
\hline
Number of layers ($\pi_t$) & \{4, 6, 8\} \\
\hline
Number of nodes ($\pi_t$) & \{200, 400, 600\} \\
\hline
Alpha $\alpha$ & \{0.5, 1.0\} \\
\hline
Beta $\beta$ & \{0.5\} \\
\hline
Batch Size & \{256, 512\} \\
\hline
Learning Rate & \{0.1, 0.01\} \\
\hline
Learning rate Decay & \{1e$^{-1}$\} \\
\hline
Learning Scheduler Step & \{10, 15\} \\
\hline
Weight Decay & \{1e$^{-4}$\} \\
\hline
Dropout & \{0.1\} \\
\hline
Activation & \{$\mathrm{Tanh}$, $\mathrm{ELU}$\} \\
\hline
\end{tabular}
\caption{Hyperparameter search range for NICE (our proposed method) and baselines on semi-synthetic datasets.}
\label{tab:hyperparameter_search}
\end{table}

In this section, we provide details of the experiments conducted to evaluate the performance of NICE against various baselines. We begin by outlining the baselines considered in the experiments, followed by a comparison of NICE performance with these baselines across different scenarios, including zero-shot tasks. Finally, we provide details of the parameters used in NICE and the baselines. In all our experiments, we use the datasets as described in the Data Simulation section and all models are evaluated using the square root of PEHE metric defined in equation~\ref{eq:epsilon-pehe-formula}.   

%Evaluating causal effects presents inherent challenges, primarily due to the difficulty in obtaining ground truth for counterfactual outcomes in observational datasets. The existing literature often addresses this by generating potential outcomes using semi-synthetic datasets. As detailed in the data simulation section, we utilize images as treatments to generate potential outcomes. Specifically, we use the Posterlens dataset \citep{aptlin2021posterlens}, considering few example movie posters shown in Figure~2,  therein as treatments. 
%%These images contain rich and diverse information, enhancing the robustness of our experiments. 
%
%It is important to note that, to the best of our knowledge, there is no prior work addressing the utilization of image treatments attributes for causal effect estimation. We evaluate our model, NICE, against multiple baselines, under various setups including zero-shot tasks, to validate its effectiveness. We conduct experiments with different number of treatments (4, 8, and 16) and use the PEHE metric defined in equation~\eqref{eq:epsilon-pehe-formula} to evaluate a model's performance. Further, we compare NICE performance under various scenarios of varying treatment assignment bias.

\subsection*{Baseline Methods}
We compare our framework with adaptations of existing methods, to our problem, that leverage treatment attributes for estimating causal effects. 
%Additionally, we validate our results against a naive approach that imputes missing potential outcomes in the observed data with mean values. 
To that end, we include modified versions of GraphITE \cite{harada2021graphite}, CaML \cite{nilforoshan2023zero}, and the Structured Intervention Network (SIN) \cite{kaddour2021causal} in the baselines, as these algorithms incorporate treatment attributes to enhance causal effect estimation. 
We also aim to include an additional baseline that does not use treatment information, to effectively demonstrate the benefits of incorporating treatment information in the causal effects estimation. As NICE primarily operates on treatment head networks, we consider TARNet~\citep{shalit2017estimating} as another baseline that does not use treatment information in the estimation. 

%We also consider TARNET~\cite{shalit2017estimating} as another baseline, though it does not use the treatment information, for effectively showing the benefit of including treatment's information on the causal effects estimates. All models are evaluated using the \(\sqrt{\epsilon \text{PEHE}}\) metric. 
%To estimate potential outcomes across multiple treatments, we employ TARNet \cite{shalit2017estimating} and evaluate our framework using the \(\sqrt{\epsilon \text{PEHE}}\) metric. 
%While many algorithms exist for estimating causal effects across multiple treatments, we specifically choose to compare NICE with methods that utilize treatment attributes. 
%Although NICE primarily operates on treatment head networks, we also conduct experiments with the TARNet algorithm to further validate our claims of improving counterfactual estimates when treatment attributes are integrated into the algorithm.

GraphITE utilizes graphs as treatments to improve causal effect estimation with the HSIC criterion, reducing bias introduced by the treatment representation space. In the performance comparison Table~\ref{tab:pehe_comparison}, we include results of our experiments for GraphITE with the HSIC criterion. Similarly, we compare the performance of our algorithm with the SIN algorithm, which primarily relies on Robinson decomposition to include a quasi-convergence guarantees for estimators. CaML (causal meta-learning) uses a meta-learning approach to estimate pseudo outcomes of estimators. One common modification required for our experiment was to replace the graph representation network in these algorithms, as they primarily utilize graphs as treatments, with an image representation network to evaluate NICE with these methods.

\subsection*{NICE Performance Assessment}
%Q1 Does incorporating Treatment attribute improve causal effect
We conducted a comprehensive evaluation of NICE across various experimental settings to assess its ITE estimation capabilities. The experimental evaluation focused on testing the hypothesis that integrating treatment attributes, particularly images, enhances the accuracy of ITE estimators. To validate this hypothesis, we employed semi-synthetic datasets as described in the data simulation section. Our results demonstrate that NICE consistently outperforms the baseline algorithms in causal estimation, particularly when dealing with 4, 8, and 16 treatment groups. Results shown in all tables are means and standard deviations of $\sqrt{\epsilon_{PEHE}}$ values computed across $10$ iterations. We use bold face to indicate the best results in the tables. As the number of treatments increases, the complexity of the problem escalates. Notably, the performance gap between NICE and the baseline methods widens as the number of treatments increases, as illustrated in Table~\ref{tab:pehe_comparison}. 

%Q2 How sesntive are the outcomes of the model with treatment embeddings used for treatment attributes 
Next, we evaluate the performance of NICE in a treatment-embedding agnostic setup using a semi-synthetic dataset generated with ResNet embeddings. To assess the robustness of our approach, we also test NICE using VGG-based image embeddings~\cite{simonyan2014very} to account for potential dataset biases that might arise from the semi-synthetic data generation process. In experiments involving variations in the number of treatments, we observe that NICE, when utilizing VGG embeddings, often surpasses the baseline methods and the ResNet-based NICE implementation, as shown in experiment evaluation Tables~\ref{tab:pehe_comparison}-\ref{tab:pehe_comparison_kappa_100}. These results underscore the model's effectiveness in causal estimation across different treatment representation embeddings.

%Q3 NICE evaluation on Zero shot
Baseline methods that we compare NICE with also claim to have zero-shot capabilities for ITE estimation on unseen treatments during training of the models. We evaluate both NICE and these baseline methods for their zero-shot abilities in the context of images as treatment setups. For this evaluation, we use a modified version of the PEHE metric, referred to as the rooted Zero-Shot PEHE metric (\(\epsilon_{\text{PEHE}}^{\text{ZS}}\)), which is defined as follows:
\begin{equation}
\label{eq:epsilon-pehe-zeros-shot-formula}
\epsilon_{\text{PEHE}}^{\text{ZS}} = \frac{1}{k-1} \sum_{\substack{a=1 \\ a \neq z}}^{k} \left[ \frac{1}{n} \sum_{i=1}^n \left(\hat{\tau}^{a, z}(\mathbf{x}_i) - \tau^{a, z}(\mathbf{x}_i)\right)^2 \right],
\end{equation}
where $z$ is the zero-shot treatment whose samples are not seen by the model during training. Observe that in the above equation, PEHE is computed using only treatment pairs that include the zero-shot treatment, as our goal is to evaluate the ITE estimation capabilities of the model in zero-shot scenarios

This metric allows us to effectively evaluate the performance of NICE and baselines in handling ITE estimation for image treatments not seen during training. As illustrated in Table~\ref{tab:pehe_comparison_zero_shot}, NICE consistently outperforms baseline methods in zero-shot ITE estimation for the image treatments.

%Q4 Effecticveness with Treatment assignment bias 
In real-world scenarios, treatment assignments can be highly skewed based on user covariates, which significantly amplifies the treatment assignment bias. To assess the performance of NICE under such conditions, we simulate scenarios by increasing the treatment assignment bias, $\kappa_a$, for a specific treatment. This ensures that treatment assignment distribution is skewed towards treatment-$a$. In particular, we consider two scenarios where $\kappa_a = 10$ for $1 \leq a <k$ and $\kappa_k = 50$ and $100.$  As illustrated in Tables~\ref{tab:pehe_comparison_kappa_50} and~\ref{tab:pehe_comparison_kappa_100}, NICE consistently outperforms baseline methods in these treatment assignment bias experiments, demonstrating its robustness in handling highly skewed treatment assignment scenarios.

%In real-world scenarios, treatment assignments can be skewed based on their efficacy and availability, which presents a significant challenge for causal inference algorithms, especially when dealing with limited information about specific treatments. 
%To assess the performance of NICE under such conditions, we modify the treatment assignment bias ($\kappa$) as discussed in the Data Simulation section, with 50 and 100 to introduce skewness into the observed dataset. As illustrated in Tables~\ref{tab:pehe_comparison_kappa_50} and~\ref{tab:pehe_comparison_kappa_100}, NICE consistently outperforms baseline methods in these treatment assignment bias experiments, demonstrating its robustness in handling skewed treatment scenarios.

%Experiment Setup
\subsection*{Model Parameters}
We briefly outline the experimental setup for optimizing NICE and baseline algorithms. For covariate representation, we use a fully connected (FC) network with $\mathrm{Tanh}$ and $\mathrm{ELU}$ activation functions. Similarly, for treatment representation and causal estimator head networks, we employ FC networks with variations in the number of nodes and layers. In dataset simulation, we generate $20,000$ instances in each experiment. The data is split into training, validation, and test sets, and performance is assessed using the PEHE metric by comparing predicted potential outcomes against the ground truth for all instances. To achieve optimal performance for NICE-ResNet, NICE-VGG, and baseline algorithms, we implement techniques such as early stopping, learning rate scheduling, weight decay, and dropout. Details of hyperparameter tuning and search ranges are provided in Table~\ref{tab:hyperparameter_search}.

%% file: conclusion.tex
\section{Conclusion}
In this study, we propose, NICE, a novel framework designed to estimate individual causal effects when images are considered as treatments. To validate the efficacy of NICE, we propose a unique semi-synthetic data simulation technique that generates potential outcomes for image treatments. NICE leverages image treatment attributes to estimate potential outcomes in scenarios involving multiple treatments. Notably, NICE demonstrates zero-shot causal effect estimation capabilities, enabling it to infer causal outcomes for novel treatments. Experimental results show that NICE consistently outperforms various baselines across different setups, achieving the best performance on the PEHE metric.
For future work, we plan to explore the scalability of NICE to more complex datasets, as well as its applicability to real-world scenarios beyond semi-synthetic simulations. Additionally, we aim to enhance the framework's interpretability and extend its capabilities to handle more diverse and complex treatment types, such as video or multimodal data.

%% file: causal-effect-estimation-for-image-interventions.bbl
\begin{thebibliography}{17}
\providecommand{\natexlab}[1]{#1}

\bibitem[{Aptlin(2021)}]{aptlin2021posterlens}
Aptlin, S. 2021.
\newblock PosterLens 25M dataset.
\newblock \emph{Kaggle}, doi: 10.34740/KAGGLE/DS/1321802.

\bibitem[{Guo, Li, and Liu(2020)}]{guo2020learning}
Guo, R.; Li, J.; and Liu, H. 2020.
\newblock Learning individual causal effects from networked observational data.
\newblock In \emph{Proceedings of the 13th international conference on web search and data mining}, 232--240.

\bibitem[{Harada and Kashima(2021)}]{harada2021graphite}
Harada, S.; and Kashima, H. 2021.
\newblock Graphite: Estimating individual effects of graph-structured treatments.
\newblock In \emph{Proceedings of the 30th ACM International Conference on Information \& Knowledge Management}, 659--668.

\bibitem[{He et~al.(2016)He, Zhang, Ren, and Sun}]{he2016deep}
He, K.; Zhang, X.; Ren, S.; and Sun, J. 2016.
\newblock Deep residual learning for image recognition.
\newblock In \emph{Proceedings of the IEEE conference on computer vision and pattern recognition}, 770--778.

\bibitem[{Johansson, Shalit, and Sontag(2016)}]{johansson2016learning}
Johansson, F.; Shalit, U.; and Sontag, D. 2016.
\newblock Learning representations for counterfactual inference.
\newblock In \emph{International conference on machine learning}, 3020--3029. PMLR.

\bibitem[{Kaddour et~al.(2021)Kaddour, Zhu, Liu, Kusner, and Silva}]{kaddour2021causal}
Kaddour, J.; Zhu, Y.; Liu, Q.; Kusner, M.~J.; and Silva, R. 2021.
\newblock Causal effect inference for structured treatments.
\newblock \emph{Advances in Neural Information Processing Systems}, 34: 24841--24854.

\bibitem[{Nilforoshan et~al.(2023)Nilforoshan, Moor, Roohani, Chen, {\v{S}}urina, Yasunaga, Oblak, and Leskovec}]{nilforoshan2023zero}
Nilforoshan, H.; Moor, M.; Roohani, Y.; Chen, Y.; {\v{S}}urina, A.; Yasunaga, M.; Oblak, S.; and Leskovec, J. 2023.
\newblock Zero-shot causal learning.
\newblock \emph{Advances in Neural Information Processing Systems}, 36: 6862--6901.

\bibitem[{Pearl(2017)}]{pearl2017detecting}
Pearl, J. 2017.
\newblock Detecting Latent Heterogeneity.
\newblock \emph{Sociological Methods \& Research}, 46(3): 370--389.

\bibitem[{Rubin(2005)}]{rubin2005causal}
Rubin, D.~B. 2005.
\newblock Causal inference using potential outcomes: Design, modeling, decisions.
\newblock \emph{Journal of the American Statistical Association}, 100(469): 322--331.

\bibitem[{Schwab et~al.(2020)Schwab, Linhardt, Bauer, Buhmann, and Karlen}]{schwab2020learning}
Schwab, P.; Linhardt, L.; Bauer, S.; Buhmann, J.~M.; and Karlen, W. 2020.
\newblock Learning counterfactual representations for estimating individual dose-response curves.
\newblock In \emph{Proceedings of the AAAI Conference on Artificial Intelligence}, volume~34, 5612--5619.

\bibitem[{Schwab, Linhardt, and Karlen(2018)}]{schwab2018perfect}
Schwab, P.; Linhardt, L.; and Karlen, W. 2018.
\newblock Perfect match: A simple method for learning representations for counterfactual inference with neural networks.
\newblock \emph{arXiv preprint arXiv:1810.00656}.

\bibitem[{Shalit, Johansson, and Sontag(2017)}]{shalit2017estimating}
Shalit, U.; Johansson, F.~D.; and Sontag, D. 2017.
\newblock Estimating individual treatment effect: generalization bounds and algorithms.
\newblock In \emph{International conference on machine learning}, 3076--3085. PMLR.

\bibitem[{Shpitser and Pearl(2012)}]{shpitser2012identification}
Shpitser, I.; and Pearl, J. 2012.
\newblock Identification of conditional interventional distributions.
\newblock \emph{arXiv preprint arXiv:1206.6876}.

\bibitem[{Simonyan and Zisserman(2014)}]{simonyan2014very}
Simonyan, K.; and Zisserman, A. 2014.
\newblock Very deep convolutional networks for large-scale image recognition.
\newblock \emph{arXiv preprint arXiv:1409.1556}.

\bibitem[{Sriperumbudur et~al.(2012)Sriperumbudur, Fukumizu, Gretton, Sch{\"o}lkopf, and Lanckriet}]{sriperumbudur2012empirical}
Sriperumbudur, B.~K.; Fukumizu, K.; Gretton, A.; Sch{\"o}lkopf, B.; and Lanckriet, G.~R. 2012.
\newblock On the empirical estimation of integral probability metrics.

\bibitem[{Thorat et~al.(2023)Thorat, Kolla, Pedanekar, and Onoe}]{thorat2023estimation}
Thorat, A.; Kolla, R.; Pedanekar, N.; and Onoe, N. 2023.
\newblock Estimation of individual causal effects in network setup for multiple treatments.
\newblock \emph{arXiv preprint arXiv:2312.11573}.

\bibitem[{Yoon, Jordon, and Van Der~Schaar(2018)}]{yoon2018ganite}
Yoon, J.; Jordon, J.; and Van Der~Schaar, M. 2018.
\newblock GANITE: Estimation of individualized treatment effects using generative adversarial nets.
\newblock In \emph{International conference on learning representations}.

\end{thebibliography}
